\documentclass{article}
\usepackage{spconf,amsmath,graphicx}

\usepackage{enumitem}
\setlist{nosep, leftmargin=14pt}

\usepackage{mwe} 

\title{Belief function-based semi-supervised learning for brain tumor segmentation }
 \name{Ling Huang$^{\dagger}$ \qquad Su Ruan$^{\star}$ \qquad Thierry Denoeux$^{\dagger}{\S }$}

 \address{$^{\dagger}$ Universit\'e de technologie de Compi\`egne, CNRS, Heudiasyc, Compi\`egne, France \\
     $^{\star}$ University of Rouen Normandy, Quantif, LITIS,Rouen, France\\
     $^{\S }$ Institut universitaire de France, Paris, France}
\begin{document}
%
\maketitle
\begin{abstract}
Precise segmentation of a lesion area is important for optimizing its treatment. Deep learning makes it possible to detect and segment a lesion field using annotated data. However, obtaining precisely annotated data is very challenging in the medical domain. Moreover, labeling uncertainty and imprecision make segmentation results unreliable. In this paper, we address the uncertain boundary problem by a new evidential neural network with an information fusion strategy, and the scarcity of annotated data by semi-supervised learning. Experimental results show that our proposal has better performance than state-of-the-art methods.
\end{abstract}
\section{Introduction}
\label{sec:intro}

Deep learning has achieved great success in many computer vision tasks with abundant labeled training data, such as image recognition, object detection and segmentation, image generation, etc. However, acquiring big labeled training data in the medical domain is particularly challenging, especially for image segmentation. Region labeling in medical image segmentation tasks requires not only careful delineation of boundaries but also qualified expertise. Many methods have been developed to address the scarse annotation problem in deep learning, such as self-training \cite{li2019transformation}, adversarial training \cite{goodfellow2014generative}, co-training \cite{peng2020deep} and clustering \cite{lian2017spatial}, in which only partially labeled or unlabeled data are used. However, training a segmentation model with only unlabeled data cannot always meet high precision requirements. Therefore, researchers now rather focus  on semi-supervised learning. Another significant problem for medical segmentation is the uncertainty and imprecision of boundary annotation. There is usually a big contradiction between the increasing demand for accurate radiotherapy and the shortage of accurate annotations. Although probabilistic approaches \cite{zhou2020brain}\cite{lapuyade2017segmenting} have shown great achievements in medical segmentation, they are limited when it comes to addressing uncertainty and imprecision problems. Belief function theory \cite{dempster1967upper} \cite{shafer1976mathematical}, a theory for modeling, fusing and reasoning with uncertain and/or imprecise information, has been successfully applied in many fields, such as image analysis, information fusion, clustering, etc. Since we only have partially labeled data, the uncertainty is increased. Thus, we propose to use belief function theory for partially supervised image segmentation. The main contributions of this paper can be summarized as follows: (1) An end-to-end evidential deep framework is proposed for brain tumor segmentation; (2) A deep semi-supervised learning method allows us to train deep network with fewer labels; (3) A fusion strategy is applied to use probabilities and belief functions to address the uncertainty boundary problem.

\section{Related work}

\subsection{Semi-supervised medical segmentation}

Techniques for semi-supervised medical image segmentation can be divided into three classes: 1) self-learning methods (using image-level labels); 2) graph-constrained methods (using bounding box information) 3) Generative adversarial neural network (GAN)-based methods (using adversarial learning). Baur et al \cite{baur2017semi} lift the concept of auxiliary manifold embedding for semi-supervised learning using Fully Convolutional Networks \cite{long2015fully} with random feature embedding. The labelled data are employed to optimize the network with a primary loss function and the training batches are further augmented with unlabelled samples and their prior maps. 
In \cite{xu2017deep}, Xu et al use a rectangle as a soft constraint by transforming it into an Euclidean distance map and predicts object mask with convolutional neural network. 
In \cite{mondal2018few}, Mondal et al design a few-shot 3D multi-modality medical image segmentation method with GAN. A $K+1$ class prediction method is used to help GAN output plausible predictions for unlabelled true data by restricting its output for fake examples. 

\subsection{Belief function theory}

Belief function theory was first introduced by Dempster \cite{dempster1967upper} and Shafer \cite{shafer1976mathematical} and was further popularized and developed by Smets \cite{smets1990combination}. The great expressive power of belief function theory allows us to represent evidence in a more faithful way than using probabilities alone. Let $\Omega =\{\omega _{1}, \omega _{2}, ..., \omega _{K}\} $ be a finite set of hypotheses about some question. Evidence about $\Omega$ can be represented by a mapping $m$ from $2^\Omega$ to $[0,1]$ such that $\sum _{A\subseteq \Omega }m(A)=1$, called a mass function. For any hypothesis $A\subseteq\Omega$, the quantity $m(A)$ represents the mass of belief allocated to $A$ and to no more specific proposition. Two mass functions $m_{1}$ and $m_{2}$ derived from two independent items of evidence can be combined by Dempster's rule \cite{shafer1976mathematical} defined as 
\begin{equation}
\label{eq:dempster}
    (m_{1}\oplus m_{2})(A)=\frac{1}{1-\kappa }\sum _{B\cap C=A}m_{1}(B)m_{2}(C),
\end{equation}
for all $A\subseteq \Omega, A\neq \emptyset$, and $(m_{1}\oplus m_{2})(\emptyset)=0$. In \eqref{eq:dempster}, $\kappa$ represents the degree conflict between $m_{1}$ and $m_{2}$ defined as
\begin{equation}
\label{eq:conflict}
    \kappa=\sum _{B\cap C=\emptyset}m_{1}(B)m_{2}(C).
\end{equation}

In \cite{denoeux2000neural}, Denoeux proposed an evidential neural network classifier. It summarizes $I$ prototypes $P$ by K-means. Each prototype $p_{i}$ is a piece of evidence supporting class $x$, whose reliability decreases with the Euclidean distance $d_{i}$ between $x$ and $p_{i}$. Mass function induced by prototype $p_{i}$ is
\begin{equation}
m_i(\omega _{k})=\alpha _{i}u_{ik}\exp(-\gamma _{i}d_{i}^{2}),
\end{equation}
$u_{ik}$ is the membership degree,$i=1,...I, k=1...,K$, $K$ is the number of class. $\alpha_{i}$ and $\gamma_{i}$ are two tuning parameters. The result for all prototypes that support certain class can be combined by the Dempster's rule.

\section{MATERIALS AND METHODS}
\label{sec:pagestyle}
\subsection{Dataset and pre-processing}
\begin{figure}[h!]
\centering
\includegraphics[scale=0.25]{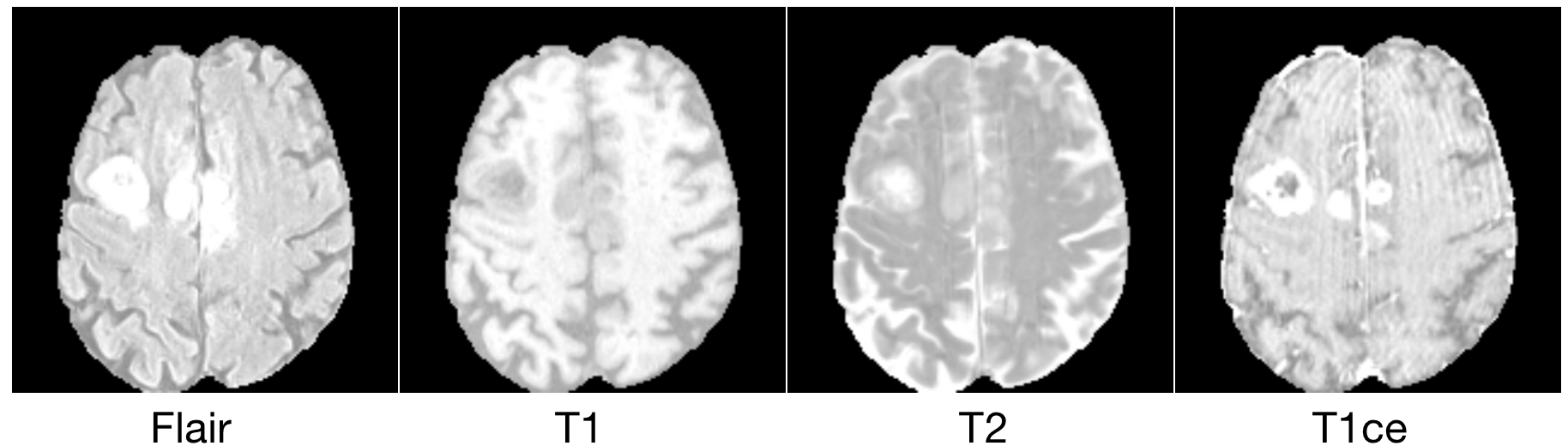}
\caption{Example of processed input images.}
\label{fig:example}
\end{figure}

 The BraTs2018 brain tumor dataset contains 285 cases of patients for training and 66 for validation. Compared with BraTs2018, BraTs2019 adds 50 additional cases; here we hold the extra 50 cases as a test set. For each patient, we have four kinds of brain MRI volumes: Flair, T1, T2, T1ce, and the tumor segmentation ground truth, as represented in Fig.~\ref{fig:example} showing an example of the four modalities. Each volume has $155\times240\times240$ voxels. In order to facilitate the training process, we first sliced 3D volumes into 2D images and normalized each modality with mean and standard deviation for brain region separately. We then center-cropped images with $160\times160$ pixels. Finally, we combined the four modality images into 3D tensors as network inputs. 

There are four labels: the GD-enhancing tumor (ET-label 4), the peritumoral edema (ED-label 2), the necrotic and non-enhancing tumor core (NCR/NET-label 1), and the background (label 0). For evaluation, there are three overlap regions: WT (Whole Tumor, ET+ED+NET), ET (enhancing tumor) and TC (tumor core, ET+NET). 

\subsection{Network Structure}
\begin{figure*}[h!]
\setlength{\belowcaptionskip}{-20pt} 
\setlength{\abovecaptionskip}{-20pt}
\centering
\includegraphics[scale=0.45]{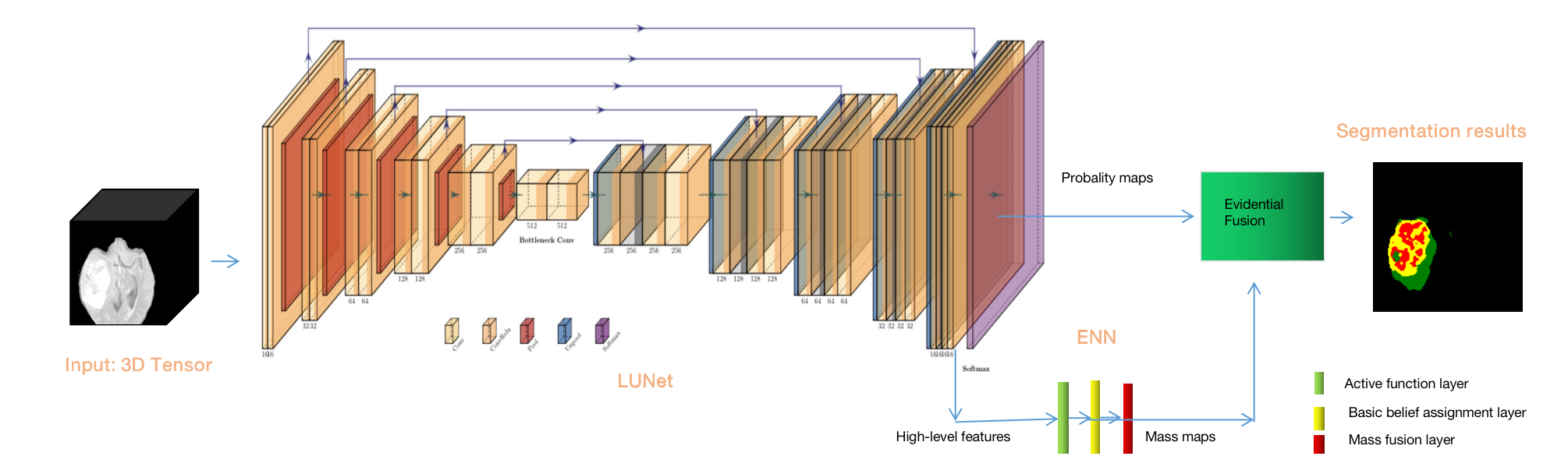}
\caption{Proposed evidential medical image segmentation framework.}
\label{fig:brats}
\end{figure*}

Fig \ref{fig:brats} shows the proposed network architecture composed of a light UNet (LUNet), an evidential neural network (ENN) \cite{denoeux2000neural} and an evidential fusion layer (EF). 
Firstly we use a modified UNet \cite{ronnebergerconvolutional} with two additional down-sampling layers: $4 \rightarrow 16$, $16\rightarrow 32$, and two additional up-sampling layers: layers: $32 \rightarrow 16$, $16 \rightarrow 4 $, and shrink the number of neuron nodes to get a light UNet. Details can be found in Fig \ref{fig:brats}. The output $p_U$ from LUNet is the vector of probabilities for pixels belongs to four class 0, 1, 2, 4. Secondly, we transfer high-level feature into ENN to better represent uncertainty. A probabilistic network may give pixels a very high probability of belonging to one class, 
even for the boundary pixels. However, a pixel at the boundary between two regions of generally should have similar probabilities of belonging to both regions due to the low contrast in the boundary. To deal with this problem, and inspired by the ENN introduced in \cite{denoeux2000neural}, we use an ENN module composed of three layers: 1) an activation layer in orange (Fig \ref{fig:brats}), which calculates distance maps between features and cluster centers; 2) a basic assignment layer in yellow, which converts activated feature values into a mass function based on distance maps; and 3) a mass fusion layer in blue for combining the contribution of each feature to reach a final decision. This ENN module outputs for each pixel five mass values: four masses corresponding to the four classes 0, 1, 2, 4 and an additional mass assigned to the ``don't know'' class $\Omega$. If some pixels are difficult to classify into one of the four classes, they are assigned to $\Omega$.
\subsection{Multi-source fusion strategy}
Since only part of the data are labeled, the uncertainty is higher than in the fully supervised case. We propose to use the belief function to deal with this problem. Probabilistic classifiers assign a probability to each class, which is not reliable for pixels close to region boundaries. In contrast, belief function classifiers compute a mass function that assigns a mass to each class and an uncommitted mass to the full set $\Omega$. Here we propose an additional evidential fusion layer to combine both segmentation results from the UNet and the ENN using the Dempster’s rule \cite{shafer1976mathematical}. Denoting the pixel-wise segmentation outputs of LUNet and ENN as, respectively, $p_U$ and $m_E$, the combined mass function is

\begin{equation}
    (p_{U}\oplus m_{E})(a_i)=\frac{1}{1-\kappa }\sum _{b\cap c=a_i}p_{U}(b)m_{E}(c),
\end{equation}
for all $a_i\subseteq \Omega=\{0,1,2,4\}, a_i\neq \emptyset$, and $(p_{U}\oplus m_{E})(\emptyset)=0$, where $\kappa$ is the degree of conflict defined in Eq. \eqref{eq:conflict}. Here, a high degree of conflict means that the two classifiers assign a pixel to different classes, which indicates high segmentation uncertainty.

\subsection{Semi-supervised learning}
Assuming that only part of the database has been labeled, we propose a semi-supervised learning method that aims to obtain a segmentation accuracy close to that of a fully-supervised learning method. For each input image $x$, we use several transformations, such as Gaussian noise, to get new images, noted $x_t$. Similar images are expected to produce similar segmentation maps even if some transformations have been performed on them. Here we train the network with 50\% of the labeled samples using the following \textsf{loss1} function, which measures the difference between the output $s_t$ and the ground truth $G$:
\begin{equation}
    \textsf{loss1}=\sum_{p=1}^{P}\sum_{t=1}^{T}1-\frac{2*\sum s^p_{t}\cap G^p}{\sum s^p_{t}+\sum G^p}+0.5 \times (s^p_{t}-G^p)^{^{2}},
\end{equation}
where $T$ is the numbers of transformed image type, $P$ is the number of pixels for segmentation outputs, and $G$ are the segmentation labels. For the 50\% of data without labels we use the following \textsf{loss2} criterion, which measures the difference between the original output $s$ and the transformed output $s_t$:
\begin{equation}
    \textsf{loss2}=\sum_{p=1}^{P}\sum_{i\neq t}^{T}1-\frac{2*\sum s^p_{i}\cap s^p_{t}}{\sum s^p_{i}+\sum s^p_{t}}+0.5 \times (s^p_{i}-s^p_{t})^{^{2}}.
\end{equation}
During training, when the iteration counter is even, \textsf{loss1} is used, and vice \textsf{loss2} is used.

\section{Experiment and results}
\label{sec:typestyle}

\subsection{Performance measures}

 Here we use Dice score, Specificity, Positive predictive value (PPV) and Hausdorff distance as performance measures. These criteria are defined as follows:
\begin{equation}
\textsf{Dice}(P,T)=\frac{2\times TP}{FP+2\times TP+FN},
\end{equation}
\begin{equation}
\textsf{PPV}(P,T)=\frac{TP}{TP+FP},
\end{equation}
\begin{equation}
\textsf{Sensitivity}(P,T)=\frac{TP}{TP+FN},
\end{equation}
where TP, FP and FN denote, respectively, the numbers of true positive, false positive and false negative samples, and
\begin{multline}
\textsf{Hausdorff} = \\
\max\left (\max_{i\in s}\min_{j\in G} d(i,j),\max_{j \in G}\min_{i \in s} d(i,j)\right ),
\end{multline}
where $d$ represents the Euclidean distance.

\subsection{Results and Analysis}

 \begin{figure*}[h]
\centering
\includegraphics[scale=0.3]{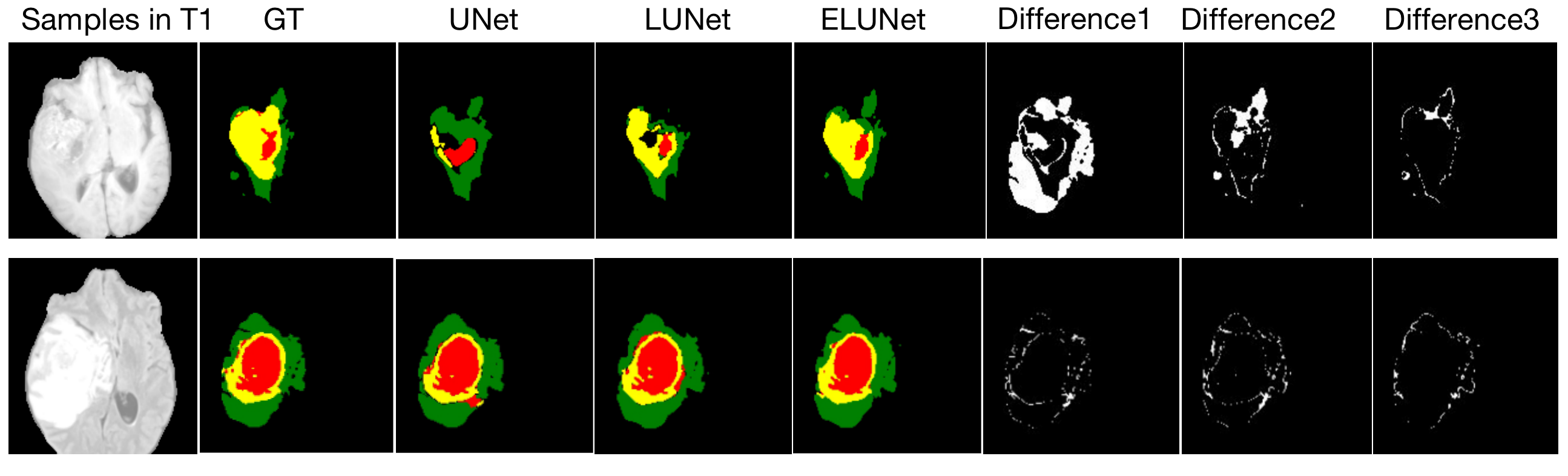}
\label{fig:comparision}
\caption{Segmentation results of brain tumor with different methods of two patients corresponding to the first row and the second one, respectively. From left to right, it shows input samples, ground truth (GT), the results of UNet, LUNet and ELUNet, the difference between GT and UNet, GT and LUNet, and GT and ELUNet.}
\label{fig:imshow}
\end{figure*}

\begin{table*}[ht]
  \centering
  \caption{Comparison of different attributes in BraTs2019 dataset}
  \label{CNN}
  \scalebox{0.80}{
  \begin{tabular}{lllllllllllllll}
  \hline
  \multicolumn{1}{l}{Methods} & \multicolumn{1}{l}{Learning} & \multicolumn{1}{l}{Parameters}  &\multicolumn{3}{c}{Dice}&\multicolumn{3}{c}{PPV}&\multicolumn{3}{c}{Sensitivity}&\multicolumn{3}{c}{Hausdorff}\\
   &&& WT&TC&ET &WT&TC&ET &WT&TC&ET& WT&TC&ET\\
     \hline
   Unet(baseline)&supervised&39.40M&77.58&88.27&90.14&83.33&91.01&93.38&89.66&94.20&93.61&1.59&0.90&0.82 \\
   LUnet&supervised&9.876M&75.52&84.14&90.35&78.95&87.56&95.04&90.40&92.53&91.53&1.71&1.07&0.82 \\

ELUnet&supervised&9.877M&\textbf{87.19}&\textbf{91.97}&\textbf{92.15}&\textbf{87.04}&\textbf{93.64}&\textbf{93.71}&\textbf{91.30}&\textbf{95.55}&\textbf{95.32}&\textbf{1.10}&\textbf{0.78}&\textbf{0.76}\\
  \hline
Unet(baseline)&semi-supervised& 39.40M&56.78&75.46&74.26&62.56&79.75&80.73&61.58&81.34&79.18&2.04&1.77&1.70\\
LUnet&semi-supervised&9.876M&82.74&86.58&76.99&88.76&91.42&99.84&89.64&91.63&94.96&1.49&0.99&1.14 \\
ELUnet&semi-supervised&9.877M&\textbf{84.01}&\textbf{89.73}&\textbf{89.94}&\textbf{88.51}&\textbf{91.62}&\textbf{91.55}&\textbf{91.47}&\textbf{94.96}&\textbf{94.54}&\textbf{1.40}&\textbf{0.86}&\textbf{0.85} \\
  \hline
  \end{tabular}}
  \label{tab:result}
\end{table*}

\begin{table}[h]
  \centering
  \caption{Comparison with the state-of-the art methods (Dice score) on BraTS 2018 validation dataset}
  \label{CNN}
  \scalebox{0.80}{
  \begin{tabular}{llllll}
  \hline
  \multicolumn{1}{l}{Methods}  &\multicolumn{3}{c}{Dice}& \multicolumn{1}{c}{Dice}\\
   & WT&TC&ET &mean\\
     \hline
ELUnet&86.16&\textbf{90.27}&\textbf{85.15} &\textbf{87.19}\\
No-new-Net \cite{isensee2018no}& 90.62& 84.54& 80.12& 85.09\\
DMFNet\cite{chen20193d}& \textbf{91.26}& 86.34&80.87&86.15\\
NVDLMED \cite{myronenko20183d} &90.68& 86.02 &81.73&86.14 \\
C-VNet\cite{sharif2020active} &	90.48&83.64&77.68&83.93\\
\hline
\end{tabular}}
\label{tab:state-of-the-art}
\end{table}
 \begin{figure}[h]
\centering
\includegraphics[scale=0.25]{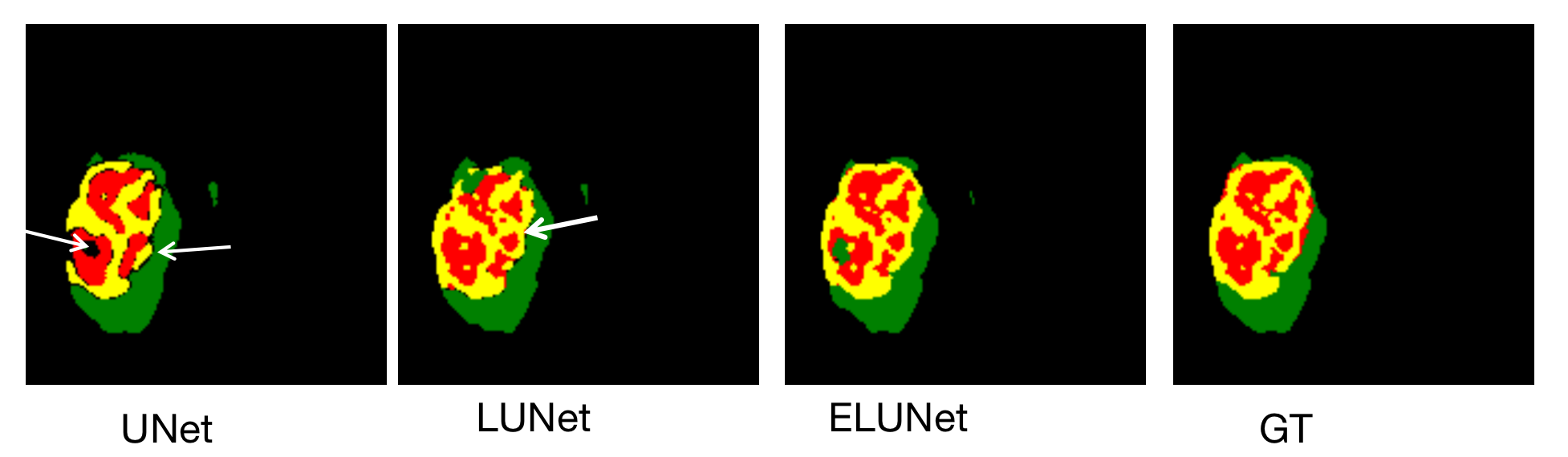}
\caption{Examples of uncertainty boundary. Green is ED, yellow is ET, red is NET. There are some black pixels between them (indicated by the white arrow in the picture), which is difficult to be assigned to ED, ET or NET.}
\label{fig:uncertainty}
\end{figure}

 The learning rate for LUnet and ENN is set to 0.001 and 0.01, respectively. Quantitative comparison are reported in Tables \ref{tab:result} and \ref{tab:state-of-the-art}. UNet is the original UNet framework, which is considered as the baseline model. LUNet is our re-designed light UNet. ELUNet is our final proposal. We can observe that semi-supervised learning does not perform well on UNet if only partially labeled training data are used. The reason is that there are more parameters to optimise, which makes the network more difficult to train. Furthermore, LUNet and ELUNet have better performance according to the four criteria compared with the baseline method UNet. We can also remark that the performance of ELUNet trained with partially annotated samples is comparable to those of fully supervised methods. 

We also compared our method with the state-of-the-art approaches in Table \ref{tab:state-of-the-art}. ELUnet achieves scores of 88.16\%, 90.27\% and 85.15\% for WT, TC and ET, respectively. Compared to the best scores achieved by DMFNet \cite{chen20193d}, ELUnet shows better performance for TC and ET with 4\% and 5\% increase, respectively. The performance for WT is not as good as that those of other methods, which may be explained by the fact that we take background loss into consideration, under the assumption that only partially annotated training samples are available.

Fig \ref{fig:imshow} shows the semi-supervised results with different methods. As can be seen from Fig \ref{fig:imshow}, LUNet and ELUnet yield better results than UNet with semi-supervised learning. When the segmentation object is clear (the yellow and red region in the first and second row, respectively), all three methods show good performance. If the region contains uncertain pixels, both UNet and LUNet have difficulties to segment them (columns 6 and 7), while ELUNet yields better results for controversial boundary segmentation thanks to ENN and EF (column 8). Fig \ref{fig:uncertainty} shows the uncertain boundary of a brain tumor. The segmentation results of UNet and LUNet are more uncertain than those of ELUNet if the pixels are close to the boundary between two classes.

\section{Conclusion}
In this work, a belief function-based medical image segmentation network, called ELUNet, has been proposed with semi-supervised learning. We first modified the Unet framework to get a light Unet. Then we transferred feature map extracted from LUnet to an ENN module for high-level semantic evidence fusion and decision. Finally, decisions from LUNet and ENN were combined by Dempster's rule to reach a final segmentation result. In the future, we plan to further investigate uncertain boundary analysis for medical images and tackle 3D volume segmentation.

\vfill
\pagebreak
\section{Compliance with Ethical Standards}
\label{sec:Compliance with Ethical Standards}

This research study was conducted retrospectively using human subject data made available in open access by ( https://www.med.upenn.edu/sbia/brats2018/data.html). Ethical approval was not required as confirmed by the license attached with the open access data.

\section{Acknowledgments}
\label{sec:acknowledgments}
This work was supported by [China Scholarships Council (No.201808331005)]. This work was carried out in the framework of the Labex MS2T, which was funded by the French Government, through the program "Investments for the future" managed by the National Agency for Research (Reference ANR-11-IDEX-0004-02)

\bibliographystyle{IEEEbib}
\bibliography{strings,refs}

\begin{thebibliography}{10}

\bibitem{li2019transformation}
X.M. Li, L.~Yu, H.~Chen, C.W. Fu, and P.A. Heng,
\newblock ``Transformation consistent self-ensembling model for semi-supervised
  medical image segmentation,''
\newblock {\em arXiv preprint arXiv:1903.00348}, 2019.

\bibitem{goodfellow2014generative}
G.~Ian, P.A. Jean, M.~Mehdi, B.~Xu, W.F. David, O.~Sherjil, C.~Aaron, and
  B.~Yoshua,
\newblock ``Generative adversarial nets,''
\newblock in {\em Advances in neural information processing systems},
  Montréal, Canada, Dec, 2014, pp. 2672--2680.

\bibitem{peng2020deep}
J.Z. Peng, E.~Guillermo, P.~Marco, and D.~Christian,
\newblock ``Deep co-training for semi-supervised image segmentation,''
\newblock {\em Pattern Recognition}, p. 107269, 2020.

\bibitem{lian2017spatial}
C.F. Lian, S.~Ruan, Den{\oe}ux T, H.~Li, and P.~Vera,
\newblock ``Spatial evidential clustering with adaptive distance metric for
  tumor segmentation in fdg-pet images,''
\newblock {\em IEEE Transactions on Biomedical Engineering}, vol. 65, no. 1,
  pp. 21--30, 2017.

\bibitem{zhou2020brain}
T.X. Zhou, C.~Stephane, V.~Pierre, and S.~Ruan,
\newblock ``Brain tumor segmentation with missing modalities via latent
  multi-source correlation representation,''
\newblock {\em arXiv preprint arXiv:2003.08870}, 2020.

\bibitem{lapuyade2017segmenting}
L.L. J{\'e}r{\^o}me, J.H. Xue, and S.~Ruan,
\newblock ``Segmenting multi-source images using hidden markov fields with
  copula-based multivariate statistical distributions,''
\newblock {\em IEEE Transactions on Image Processing}, vol. 26, no. 7, pp.
  3187--3195, 2017.

\bibitem{dempster1967upper}
A.~P. Dempster,
\newblock ``Upper and lower probability inferences based on a sample from a
  finite univariate population,''
\newblock {\em Biometrika}, vol. 54, no. 3-4, pp. 515--528, 1967.

\bibitem{shafer1976mathematical}
G.~Shafer,
\newblock {\em A mathematical theory of evidence}, vol.~42,
\newblock Princeton university press, 1976.

\bibitem{baur2017semi}
C.~Baur, S.~Albarqouni, and N.~Navab,
\newblock ``Semi-supervised deep learning for fully convolutional networks,''
\newblock in {\em International Conference on Medical Image Computing and
  Computer-Assisted Intervention}, Quebec, Canada, Sep, 2017, Springer, pp.
  311--319.

\bibitem{long2015fully}
L.~Jonathan, S.~Evan, and D.~Trevor,
\newblock ``Fully convolutional networks for semantic segmentation,''
\newblock in {\em Proceedings of the IEEE conference on computer vision and
  pattern recognition}, Boston, USA, Jun, 2015, pp. 3431--3440.

\bibitem{xu2017deep}
N.~Xu, B.~Price, S.~Cohen, J.~Yang, and T.~Huang,
\newblock ``Deep grabcut for object selection,''
\newblock {\em In BMVC}, 2017.

\bibitem{mondal2018few}
A.K. Mondal, J.~Dolz, and C.~Desrosiers,
\newblock ``Few-shot 3{D} multi-modal medical image segmentation using
  generative adversarial learning,''
\newblock {\em arXiv preprint arXiv:1810.12241}, 2018.

\bibitem{smets1990combination}
Ph. Smets,
\newblock ``The combination of evidence in the transferable belief model,''
\newblock {\em IEEE Transactions on pattern analysis and machine intelligence},
  vol. 12, no. 5, pp. 447--458, 1990.

\bibitem{denoeux2000neural}
T.~Denoeux,
\newblock ``A neural network classifier based on {Dempster-Shafer} theory,''
\newblock {\em IEEE Transactions on Systems, Man, and Cybernetics-Part A:
  Systems and Humans}, vol. 30, no. 2, pp. 131--150, 2000.

\bibitem{ronnebergerconvolutional}
O.~Ronneberger, P.~Fischer, and TU~net. Brox,
\newblock ``Convolutional networks for biomedical image segmentation,''
\newblock in {\em Paper presented at: International Conference on Medical Image
  Computing and Computer-Assisted Intervention}, Munich, Germany, Oct, 2015.

\bibitem{isensee2018no}
F.~Isensee, P.~Kickingereder, W.~Wick, M.~Bendszus, and K.H. Maier-Hein,
\newblock ``No new-net,''
\newblock in {\em International MICCAI Brainlesion Workshop}. Springer, 2018,
  pp. 234--244.

\bibitem{chen20193d}
C.~Chen, X.P. Liu, M.~Ding, J.F. Zheng, and J.Y. Li,
\newblock ``3d dilated multi-fiber network for real-time brain tumor
  segmentation in mri,''
\newblock in {\em International Conference on Medical Image Computing and
  Computer-Assisted Intervention}. Springer, 2019, pp. 184--192.

\bibitem{myronenko20183d}
A.~Myronenko,
\newblock ``3d mri brain tumor segmentation using autoencoder regularization,''
\newblock in {\em International MICCAI Brainlesion Workshop}. Springer, 2018,
  pp. 311--320.

\bibitem{sharif2020active}
M.I. Sharif, J.P. Li, M.A. Khan, and M.A. Saleem,
\newblock ``Active deep neural network features selection for segmentation and
  recognition of brain tumors using mri images,''
\newblock {\em Pattern Recognition Letters}, vol. 129, pp. 181--189, 2020.

\end{thebibliography}

\end{document}